\documentclass[10pt, letterpaper]{article}  

\usepackage[round,compress]{natbib}
\usepackage{style}

\newcommand\numberthis{\addtocounter{equation}{1}\tag{\theequation}}  

\newcommand{\ignore}[1]{}

\title{
\tbf{Efficient Covariance Estimation from Temporal Data}}
\date{}

\author{
\small Hrayr Harutyunyan$^1$, Daniel Moyer$^1$, Hrant Khachatrian$^2$, Greg Ver Steeg$^1$, and~Aram Galstyan$^1$\\[0.1in]
\small $^{1}$Information Sciences Institute, University of Southern California\\
\small $^{2}$YerevaNN Research Lab\\
\small E-mails: \href{mailto:hrayrh@isi.edu}{\{{hrayrh, gregv, galstyan}\}@isi.edu}, \href{mailto:moyerd@usc.edu}{moyerd@usc.edu}, \href{mailto:hrant@yerevann.com}{hrant@yerevann.com}
}

\usepackage{tikz}
\usetikzlibrary{arrows.meta,calc,decorations.markings,math,arrows.meta}

\begin{document}
\maketitle

\begin{abstract}
Estimating the covariance structure of multivariate time series is a fundamental problem with a wide-range of real-world applications from financial modeling to fMRI analysis.
Despite significant recent advances, current state-of-the-art methods are still severely limited in terms of scalability, and do not work well in high-dimensional undersampled regimes.
In this work we propose a novel method called T-CorEx, that scales to very large temporal datasets that are not tractable with existing methods and gives state-of-the-art results in highly undersampled regimes.
T-CorEx optimizes an information-theoretic objective function to learn an approximately modular latent factor model for each time period and applies two regularization techniques to induce temporal consistency of estimates.
We perform extensive evaluation of T-Corex using both synthetic and real-world data and demonstrate that it can be used for detecting sudden changes in the underlying covariance matrix, capturing transient correlations and analyzing extremely high-dimensional complex multivariate time series such as high-resolution fMRI data. 

\end{abstract}

\section{Introduction}
Many complex systems in finance, biology, and social sciences can be modeled by multivariate time series.
One of the first steps in analyzing such time-varying complex systems is temporal covariance estimation---that is,  estimation of the covariance matrix of the variables at different time periods.
Such an analysis can be helpful for understanding relationships between components of the system, spotting trends, making predictions, detecting shifts in the underlying structure and for other tasks \citep{engle2017large, de2018factor, ahmed2009recovering, monti2014estimating}.

There is an increasing need for efficient temporal covariance estimation methods that can work in high-dimensional undersampled regimes.
In this regime, even \textit{static} covariance estimation is a formidable problem~\citep{fan2016overview}.
Extending covariance estimation to the temporal case adds unique challenges.
First, the samples from different time steps generally are not independent and identically distributed.
Second, the dynamics of the system can be quite complex (e.g., financial time series, biological systems) and hard to model without strong assumptions.
Furthermore, the number of time steps whose observations are relevant for estimating the covariance matrix at a particular time period can be small when the underlying covariance matrix changes quickly.



Current state-of-the-art temporal covariance estimation methods learn a sparse graphical model for each time period using variants of graphical lasso, while adding a regularization term for inducing temporal consistency of estimates \citep{hallac2017network, TomasiLTGL}.
Unfortunately, these methods have at least cubic time complexity and quadratic memory complexity in the number of variables, and do not scale to truly high-dimensional problems (e.g., the approaches described in Refs.~\citep{hallac2017network, TomasiLTGL} do not scale beyond thousands of time series).

As our main contribution of this work, we propose a novel temporal covariance estimation method, called temporal correlation explanation (T-CorEx), that addresses the aforementioned challenges.
T-CorEx is based on linear CorEx, which learns approximately modular static latent factor models by minimizing an information-theoretic objective function~\citep{steeg2017low}.
Given multivariate time series divided into time periods, our method trains a linear CorEx for each time period and employs two regularization techniques to enforce temporal consistency of learned models. 
T-CorEx has linear stepwise computational complexity with respect to the number of variables and can be applied to temporal data with several orders of magnitude more variables than what existing methods can handle (e.g., it takes less than an hour on a moderate PC to estimate the covariance structure for time series with $10^5$ variables). 
To our knowledge, the proposed method is the only temporal covariance estimation method that has linear time and memory complexity with respect to the number of variables.
The only assumption T-CorEx makes about the dynamics of the system is that on average, the underlying covariance matrix changes slowly with time.

We compare T-CorEx against other methods on both synthetic and real-world datasets.
Our experiments show that T-CorEx yields state-of-the-art performance in highly undersampled regimes.
More specifically, T-CorEx outperforms other methods in terms of describing existing correlations in the data (quantified by log-likelihood).
With experiments on stock market data, we demonstrate that T-CorEx can detect transient correlations and sudden change-points of the underlying covariance matrix, not detectable with static methods.
Additionally, we apply T-CorEx on high-resolution functional magnetic resonance image (fMRI) data, and show that it successfully scales for multivariate time series with 150K variables and finds meaningful functional connectivity patterns.
Our implementation of T-CorEx and the source code of experiments are publicly available at \url{https://github.com/hrayrhar/T-CorEx}.

\section{Methods}\label{sec:method}
\begin{figure}
    \centering
    \includegraphics[width=0.65\textwidth]{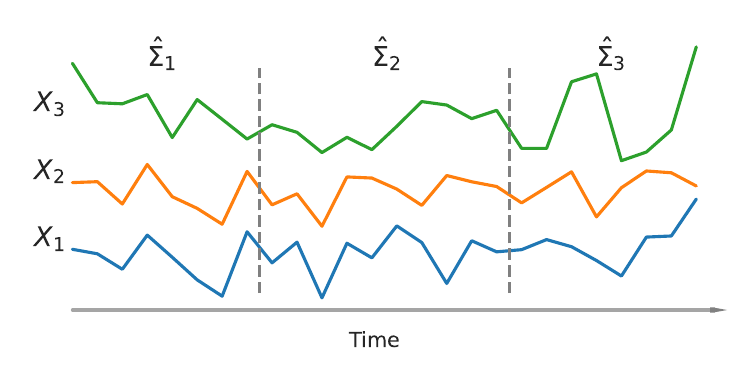}
    \caption{An example of multivariate time series with time-varying covariance matrix, divided into 3 time periods. Temporal covariance estimation methods need to produce a covariance estimate for each time period.}
    \label{fig:timeseries}
\end{figure}
To derive our method, we formulate the temporal covariance estimation problem for a sequence of multivariate Gaussian observations. We are given a sequence of observations ${\lbrace D_t \rbrace}_{t=1}^T$, where $D_t \in \mathbb{R}^{s_t \times p}$ is a collection of $s_t$ i.i.d. samples generated from a $p$-dimensional normal distribution with zero mean and covariance matrix $\Sigma_t \in \mathbb{S}^{p}_{++}$. The goal is to estimate the unknown $\Sigma_t$ covariance matrices for each $t$.
Without further assumptions, the problem is equivalent to having $T$ independent static covariance estimation tasks.
Usually covariance matrices of different time steps are related, and efficient temporal covariance estimation methods should exploit the relation.
However, the relation can be complicated and hard to model without strong assumptions. To avoid making such assumptions, we only assume that on average, covariance matrices of neighboring time steps are close to each other with respect to some metric. 

In practice, we often encounter regular multivariate time series (i.e., when $s_t=1$ for all $t$) and our goal is to estimate the covariance matrix for each time period of length $w$, rather than for each time step.
To do this, we divide the period $[1,\ldots,T]$ into periods of size $w$ and treat the samples of a period as i.i.d. samples (see Fig.~\ref{fig:timeseries}).
We get a temporal covariance estimation instance where the number of time periods is approximately $T/w$ and the number of samples in each time period is approximately $w$.

Any static covariance estimation method $\mathcal{M}$ can be used for temporal covariance estimation by applying it on each time period independently.
When $w$ is significantly smaller than the sample complexity of $\mathcal{M}$, this approach may produce inaccurate estimates.
A better approach is to apply $\mathcal{M}$ on each time period, but enforce temporal consistency of estimates with a regularization term~\citep{hallac2017network, TomasiLTGL}.
We use this approach to extend linear CorEx \citep{steeg2017low} for temporal covariance estimation.
There are several reasons why we choose to base our approach on this static method.
First, linear CorEx has been shown to have low sample complexity.
Second, it scales to extremely high-dimensional data (where $p$ is greater than $10^5$).
Lastly, linear CorEx exhibits blessing of dimensionality -- the number of samples required to accurately recover structure goes down as the number of observed variables increases~\citep{steeg2017low}.
These facts make linear CorEx a good candidate for temporal covariance estimation when $w \ll p$.

\subsection{Linear CorEx}\label{subsec:corex}
To proceed further we describe the notation, present some details of linear CorEx and one definition from~\citep{steeg2017low}.

\textbf{Notation.} We denote random variables with capital letters and use corresponding lowercase letters for values.
Throughout the paper we have that $X = (X_1, X_2, \ldots, X_p)$ is a $p$-dimensional random variable and $Z=(Z_1,Z_2,\ldots,Z_m)$ is an $m$-dimensional random variable. The former is used for denoting the observed variables, while the latter is used for denoting the latent factors.
$A_{1:T}$ is a shorthand for the set $\{A_1, \ldots, A_T\}$.
Throughout the paper we use several information-theoretic concepts.
The differential entropy of a random variable $X$ is denoted with $H(X)$ and is defined as $H(X) = -\mathbb{E}\left[\log p(x)\right]$.
Entropy measures uncertainty and evaluates to zero when the random variable takes a constant value almost surely.
The total correlation~\citep{watanabe} of $X$ is denoted with $TC(X)$ and is equal to $\sum_{i=1}^p H(X_i) - H(X)$.
It quantifies the dependency among a set of random variables and is zero if and only if the variables are independent.
For random variables $Z$ and $X$, the total correlation of $Z$ conditioned on $X$ is defined as $TC(Z \mid X) = \mathbb{E}_{x}\left[ TC(Z \mid X = x)\right]$.
More information on these information-theoretic quantities can be found in \citep{cover}.

\begin{figure}
\centering
\scalebox{1.0}{
\begin{tikzpicture}
    \draw (-1.5,1.8) circle (0.5cm) node {$Z_1$};
    \draw (+1.5,1.8) circle (0.5cm) node {$Z_m$};
    \draw (0, 1.8) node {$\ldots$};
    \draw[fill=gray!20] (-2.25,0) circle (0.5cm) node {$X_1$};
    \draw[fill=gray!20] (-0.75,0) circle (0.5cm) node {$X_2$};
    \draw[fill=gray!20] (+0.75,0) circle (0.5cm) node {$X_{\ldots}$};
    \draw[fill=gray!20] (+2.25,0) circle (0.5cm) node {$X_p$};
    \draw[-{Latex[scale=1.5]}] (-1.7,1.34) -- (-2.3,0.5);
    \draw[-{Latex[scale=1.5]}] (-1.4,1.31) -- (-0.85,0.49);
    
    \draw[-{Latex[scale=1.5]}] (1.4,1.31) -- (+0.85,0.49);
    \draw[-{Latex[scale=1.5]}] (1.7,1.34) -- (+2.25,0.5);
    \draw (0, -0.8) node {$\Downarrow$ (for any distribution) \hspace{0.2cm} $\Uparrow$ (for Gaussians)};
    \normalsize
    \draw (0, -1.4) node {$TC(X \mid Z) + TC(Z) = 0, \ \& \ \forall i, TC(Z \mid X_i) = 0$};
    \end{tikzpicture}
}
\caption{The directed probabilistic graphical model corresponding to a modular latent factor model.
The figure is taken from \citep{steeg2017low}.}
\label{fig:schematic}
\end{figure}
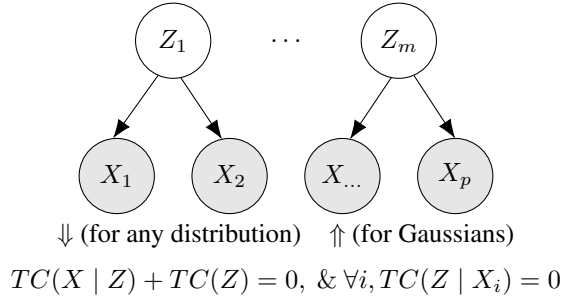

\begin{definition}
A joint distribution $p(x,z)$ with $p$ observed variables $X_{1:p}$ and $m$ hidden variables $Z_{1:m}$ is called modular latent factor model if it factorizes in the following way:
\begin{equation*}
    \forall x, z, ~~p(x,z) = \left( \prod_{i=1}^p{p(x_i\mid z_{\pi_i})}\right) \left( \prod_{j=1}^m{p(z_j)} \right),
\end{equation*}
with $\pi_i \in \{1, \ldots, m\}$.
\end{definition}
The factorization of modular latent factor models corresponds to the directed probabilistic graphical model shown in Fig.~\ref{fig:schematic}.
Modular models are easy to interpret as they provide a clustering of observed variables.

Linear CorEx~\citep{steeg2017low} is a latent factor modeling approach for estimating multivariate Gaussian distributions.
For a given $p$-dimensional Gaussian random variable $X = (X_1, \ldots, X_p)$, the algorithm finds an $m$-dimensional Gaussian random variable ${Z=(Z_1, \ldots, Z_m)}$, such that $p(x, z)$ is close to being modular.
The authors show that a jointly Gaussian distribution $p(x,z)$ is a modular latent factor model if and only if $TC(X\mid Z) + TC(Z) = 0$ and $\forall i, TC(Z\mid X_i)=0$.
Given this fact, linear CorEx parametrizes $p_W(z\mid x)=\prod_{j=1}^m \mathcal{N}(z_j; w_j^T x, 1)$ with $W\in\mathbb{R}^{m\times p}$ and solves the following optimization problem:
\begin{equation}
    \min_{W}  TC(X\mid Z) + TC(Z) + \sum_{i=1}^p Q_i,
    \label{eq:main_opt_problem}
\end{equation}
where $Q_i$ are regularization terms encouraging modular solutions (i.e. solutions with small $TC(Z\mid X_i)$).
Assuming $\forall i, \mathbb{E}[X_i]=0 \text{ and } \mathbb{E}[X_i^2] = 1$, after some modification of the original optimization problem (\ref{eq:main_opt_problem}), the algorithm solves the following problem:
\begin{equation}
\min_{W}{\sum\limits_{i=1}^{p}{\frac{1}{2}\log{\mathbb{E}[(X_i - \nu_{X_i\mid Z})^2 ]}}} + \sum_{j=1}^{m}{\frac{1}{2}\log{\mathbb{E}[Z_j^2]}},
\label{eq:final_opt_problem}
\end{equation}
where ${ \nu_{X_i\mid Z} }$ is the conditional mean of $X_i$ given $Z$ under the constraint that ${ \forall i, TC(Z\mid X_i) = 0 }$, and is calculated in the following way:
\begin{align*}
&\nu_{X_i\mid Z} = \frac{1}{1 + r_i}\sum_{j=1}^m{B_{j,i}\frac{Z_j}{\sqrt{\mathbb{E}[Z_j^2]}}},\ \ R_{j,i} = \frac{\mathbb{E}[X_iZ_j]}{\sqrt{\mathbb{E}[X_i^2]\mathbb{E}[Z_j^2]}}\\
&B_{j,i} = \frac{R_{j,i}}{1 - R_{j,i}^2}, \ \ \ r_i = \sum_{j=1}^m{R_{j,i}B_{j,i}}.
\end{align*}
After inferring the parameters $W$, the covariance matrix $\Sigma$ of $X$ is estimated with the following formula:
\begin{equation}
    \widehat{\Sigma}_{i, k \neq i} = \frac{(B^TB)_{i,k}}{(1 + r_i)(1 + r_k)} \numberthis , \ \ \
\widehat{\Sigma}_{i, i} = 1.
\label{eq:cov_formula}
\end{equation}
Note that $\widehat{\Sigma}$ can be written as a diagonal plus low-rank matrix of form $D + U^TU$, where $D \in \mathbb{R}^{p \times p}$ is a diagonal matrix and $U \in \mathbb{R}^{m \times p}$.

Although linear CorEx is derived for Gaussian random variables,
experiments show that it can be successfully applied to non-Gaussian cases.
Furthermore, we want to emphasize that linear CorEx does not model the data with a modular latent factor model (which is a fairly restricted one).
Instead, it models the data with an approximately modular model (i.e., when $TC(Z), TC(Z\mid X) \text{ and } TC(Z \mid X_i)$ are small).

\subsection{Time-Varying Linear CorEx}
As mentioned earlier, we train one linear CorEx for each time period and use regularization techniques to enforce temporal consistency of estimates (i.e., enforcing adjacent time periods to have similar covariance matrix estimates).
As a first attempt towards building time-varying linear CorEx, we consider the following optimization problem:
\begin{equation}
\min_{W_1, \ldots, W_T}{\sum_{t=1}^{T}{\mathcal{O}(W_t, D_t)} + \lambda \sum_{t=1}^{T-1} \Phi(W_{t+1} - W_{t})},
\label{eq:tcorex-simple}
\end{equation}
where $\mathcal{O}(W_t, D_t)$ is the objective of linear CorEx with parameters $W_t$ on data $D_t$, and $\Phi$ is the penalty function, which in this work is either $\ell_1$ or $\ell_2$ vector norm. The former is suitable for systems with sudden changes, while the latter is better suited for smoothly varying systems. We name the method of Eq. (\ref{eq:tcorex-simple}) \textit{T-CorEx-simple}.

While T-CorEx-simple follows the general framework of building a temporal covariance estimation method from a static covariance estimation method, it is not powerful enough to get significantly better performance compared to linear CorEx applied to time periods independently.
This happens because in linear CorEx $\widehat{\Sigma}_t$ is a function of $W_t$ and $D_t$ (see Eq. \ref{eq:cov_formula}).
Even when the regularization coefficient $\lambda$ is infinity, making $W_1=W_2=\ldots=W_T$, still we get that in general $\widehat{\Sigma}_t \neq \widehat{\Sigma}_{t+1}$ as $D_t \neq D_{t+1}$.
Therefore, the method can never produce very close estimates of covariance matrices. If we choose to explicitly penalize the difference $(\widehat{\Sigma}_t - \widehat{\Sigma}_{t+1})$, instead of the difference $(W_t - W_{t+1})$, in general, there will be no $W_1, \ldots, W_T$ that makes the new regularization term equal to zero.

To make the regularization effective, we note that $\widehat{\Sigma}_t$ depends on $D_t$ through $R_{j,i}$, which in turn depends on $\mathbb{E}[X_i Z_j]$ and $\mathbb{E}[Z_j^2]$, where the expectations are taken over samples of $D_t$ using parameters $W_t$. When $D_t$ contains a small number of samples, these quantities can be quite noisy, resulting in noisy estimates $\widehat{\Sigma}_t$. To reduce the noise without increasing the period size $w$, we propose a simple remedy. To estimate $\mathbb{E}[X_i Z_j]$ and $\mathbb{E}[Z_j^2]$ for time period $t$, we use not only the samples of $D_t$, but also samples of other time periods $D_{\tau \neq t}$. Of course, samples belonging to time periods far from $t$ are less important than the samples belonging to time periods close to $t$.
We introduce a new hyper-parameter $\beta$ that defines the rate of decay of ``importance'' as we go far from the current time period $t$.
Specifically, the samples of time period $\tau$ have weight $\alpha_t(\tau) = \beta ^ {|t - \tau|} (0 < \beta < 1)$ when they are used to estimate $\mathbb{E}[X_i Z_j]$ and $\mathbb{E}[Z_j^2]$ for time period $t$.
We estimate $\mathbb{E}[Z_j^2]$ for time period $t$ the following way:
{\normalsize
\begin{enumerate}
\item ${ Z^{(\tau)} \leftarrow D_{\tau} W_t^T + \mathcal{E}^{(\tau)}, }$\quad ${ \mathcal{E}^{(\tau)}_{\ell,j} \overset{\text{iid}}{\sim} \mathcal{N}(0, 1), }$\\
${\tau = 1, \ldots, T, }$ \quad ${\ell = 1, \ldots, s_{\tau},\quad j=1,\ldots,m }$
\item $ \widehat{\mathbb{E}[Z_j^2]} \leftarrow \left(\sum\limits_{\tau=1}^T{\alpha_t(\tau) \sum\limits_{\ell=1}^{s_{\tau}}{(Z^{(\tau)}_{\ell,j})^2}}\right) /\left(\sum\limits_{\tau=1}^T{\alpha_t(\tau) s_{\tau}}\right) .$
\end{enumerate}
}
The $\mathbb{E}[X_i Z_j]$ is estimated in a similar way.
For computational efficiency, at time period $t$ we do not consider time periods $\tau$ for which $\alpha_t(\tau) < 10^{-9}$.
Summing up, we get the following optimization problem:
\begin{equation}
\min_{W_{1:T}}{\sum_{t=1}^{T}{\tilde{\mathcal{O}}(W_t, D_{1:T})} + \lambda \sum_{t=1}^{T-1} \Phi(W_{t+1} - W_{t})},
\label{eq:tcorex}
\end{equation}
where $\tilde{\mathcal{O}}(W_t, D_{1:T})$ is the linear CorEx objective with parameters $W_t$  applied on data $D_1, D_2, \ldots, D_T$, where samples of $D_{\tau}$ have weight $\alpha_t(\tau)$. We name the method of Eq.~(\ref{eq:tcorex}) \textit{T-CorEx}.


\begin{figure*}
\begin{algorithm}[H]
\caption{T-CorEx: temporal correlation explanation. A PyTorch implementation is available at \url{https://github.com/hrayrhar/T-CorEx}.}
\begin{algorithmic}[1]
 \STATE \textbf{Input:} temporal data of form $\{D_1,\ldots, D_T\}$, where $D_t\in \mathbb{R}^{s_t \times p}$.
 \STATE \textbf{Hyperparameters:} number of latent variables $m$, regularization coefficient $\lambda$, importance decay rate $\beta$.
 \STATE \textbf{Output:} {weight matrices $W_1,\ldots,W_T$ optimizing Eq.~(\ref{eq:tcorex}) and covariance estimates $\widehat{\Sigma}_1, \ldots, \widehat{\Sigma}_T$}.
 \\
 \STATE Initialize all $W_t$ with weights of linear CorEx trained on the concatenation of $D_1,\ldots,D_T$.
 \\
 \FOR {$t = 1..T$}
    \STATE Compute mean and variance for each $X_i$ at time period $t$ using data $D_{1:T}$ with sample weights $\alpha_t(\tau)$.
    \STATE Standardize each variable of $D_t$ using the computed mean and variance.
 \ENDFOR
 \\
 \FOR{$\epsilon$ in $[0.6, 0.6^2, 0.6^3, 0.6^4, 0.6^5, 0.6^6, 0]$}
 \REPEAT
    \STATE $\tilde{D}_t \leftarrow \sqrt{1 - \epsilon^2} D_t + \epsilon \mathcal{E}$, where $\mathcal{E}_{\ell,i} \overset{\text{iid}}{\sim} \mathcal{N}(0, 1)$, for each $t=1,\ldots,T$ (with different $\mathcal{E}$ each time).
    \FOR{$t=1..T$}
        \STATE Estimate $\mathbb{E}\left[Z^2_j\right]$ and $\mathbb{E}\left[X_i Z_j\right]$ at time period $t$ using data $\tilde{D}_{1:T}$ with sample weights $\alpha_t(\tau)$.
        \STATE Compute quantities $R_{j,i}$, $r_i$, $B_{j,i}$ and $\nu_{X_i|Z}$.
        \STATE Compute the objective of the linear CorEx at time $t$ using Eq.~(\ref{eq:final_opt_problem}) and denote it with $\tilde{\mathcal{O}}(W_t,\tilde{D}_{1:T})$.
    \ENDFOR
    \STATE Form the final objective: $\mathcal{L}(W_{1:T}) \triangleq \sum_{t=1}^{T}{\tilde{\mathcal{O}}(W_t, \tilde{D}_{1:T})} + \lambda \sum_{t=1}^{T-1} \Phi(W_{t+1} - W_{t})$.
    \STATE Do one step of ADAM optimizer to update $W_{1:T}$ using $\nabla_{W_{1:T}} \mathcal{L}(W_{1:T})$.
  \UNTIL{convergence or maximum number of iterations is reached}
\ENDFOR
\\
\STATE Write $\widehat{\Sigma}_t$ in form $D_t + U_t^T U$ using Eq.~(\ref{eq:cov_formula}).
\end{algorithmic}
\end{algorithm}
\label{alg:tcorex}
\end{figure*}

\textbf{Implementation.}
We implement the objective of Eq.~(\ref{eq:tcorex}) in PyTorch and optimize it using gradient descent.
Specifically, in all our experiments we use the Adam optimizer \citep{kingma2014adam} with $\alpha=10^{-3}$ and $\beta_1=0.9$.
We initialize the weights of T-CorEx with the weights of a linear CorEx trained on all samples of all time periods.
Additionally, similar to the training of linear CorEx we use annealing rounds.
At each round we pick a noise amount $\epsilon \in [0,1]$ and replace the data with its noisy version: $\tilde{D}_t = \sqrt{1 - \epsilon^2} D_t + \epsilon \mathcal{E}$, where $\mathcal{E}_{\ell,i}  \overset{\text{iid}}{\sim} \mathcal{N}(0, 1)$.
This weakens the correlations between variables and facilitates the optimization.
We do 7 annealing rounds with $\epsilon$ schedule $[0.6^1, 0.6^2,\ldots, 0.6^6, 0]$.
The complete algorithm is presented in Alg.~1.
Our implementation is available at \url{https://github.com/harhro94/T-CorEx}.
As training of T-CorEx involves many matrix multiplications, GPUs can be used for speeding up the training.
In fact, when $p \ge 10^4$, a single GPU can reduce the training time more than 10 times.

\textbf{Computational Complexity.}\label{subsec:complexity}
The stepwise time complexity of the linear CorEx at time period $t$ is $O(n_t mp)$, where $n_t$ is the number of samples used for estimating $\mathbb{E}[X_i Z_j]$ and $\mathbb{E}[Z_j^2]$ at time period $t$ .
Since we ignore time periods $\tau$, for which $\alpha_t(\tau) < 10^{-9}$, we get that $\sum_{t=1}^T n_t = O(\sum_{t=1}^T {s_t} / \log(1/\beta))$. Therefore, the time complexity of each optimization step of T-CorEx is $O(nmp / \log(1/ \beta))$, where $n = \sum_{t=1}^T{s_t}$ is the total number of samples.
The memory complexity of T-CorEx is $O((mT + n)p)$.
Computing a covariance matrix estimate explicitly using Eq.~(\ref{eq:cov_formula}) has $O(mp^2)$ complexity.
However, one can use the fact that covariance estimates produced by T-CorEx are diagonal plus low-rank matrices and do the needed operations in $O(m^3 + m^2p)$ time without constructing $p\times p$ matrices explicitly.
For example, we can use the factorization to multiply such a matrix with other matrices efficiently.
One can compute the determinant of such matrices in $O(m^3 + m^2p)$ time
using the generalization of the matrix determinant lemma.
Furthermore, the inverse of such matrices has diagonal plus low-rank form
and can be computed in $O(m^3 +m^2p)$ time using the Woodbury matrix identity.
For more details on fast operations with diagonal plus low-rank matrices please refer to the Sec.~\ref{sec:app-fast-operations} of the appendix.

\begin{figure}
    \centering
    \includegraphics[width=0.65\textwidth]{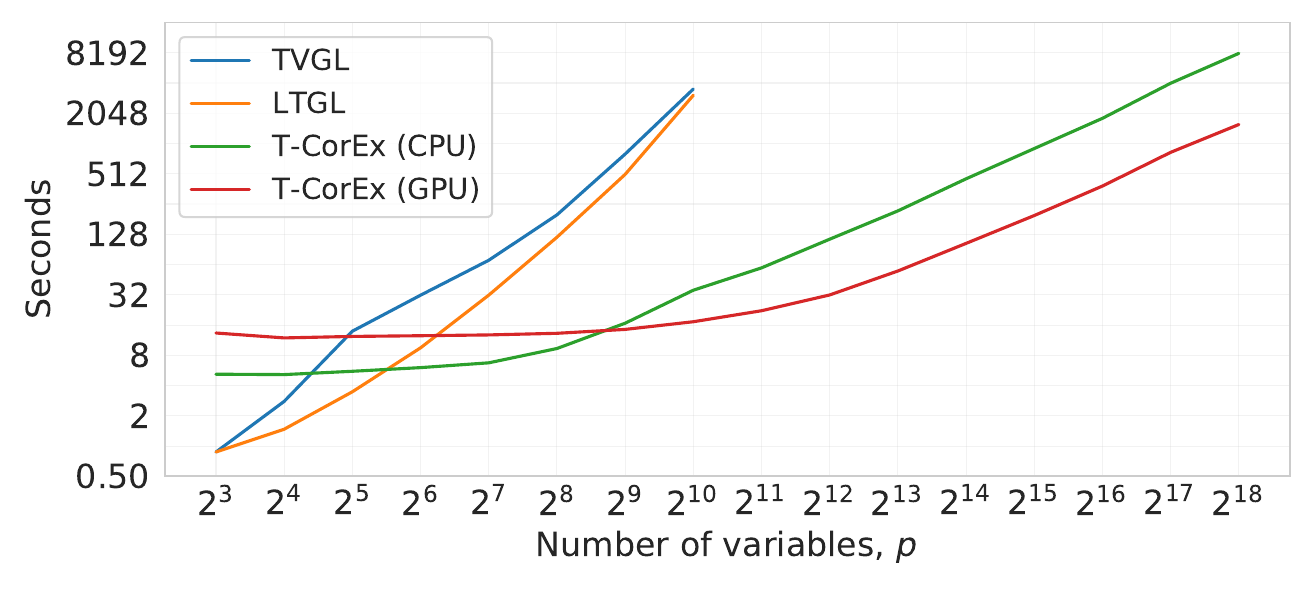}
    \caption{Scaling comparison of time-varying graphical lasso (TVGL), latent variable time-varying graphical lasso (LTGL) and T-CorEx.
    The data has $T=10$ time periods, each having 16 samples generated from a modular latent factor model with $p$ observed variables and $m=max(1, p/16)$ hidden variables.
    TVGL gives memory error starting from $p=2^{11}$, while LTGL starts to exceed the time limit of 2 hours.
    T-CorEx is trained with 64 latent factors.
    All models were trained on a computer with an Intel i5-6600K CPU, 32GB RAM, and a GeForce GTX 1080 Ti GPU.
    }
    \label{fig:scalability}
\end{figure}

We want to emphasize that the proposed method has linear stepwise computational complexity w.r.t. the number of variables $p$, assuming $n$, $T$ and $m$ do not depend on $p$. 
To our best knowledge, T-CorEx is the only temporal covariance estimation method that scales linearly with $p$. 
Time-varying graphical lasso~\citep{hallac2017network} and latent variable time-varying graphical lasso~\citep{TomasiLTGL}, which are the direct competitors of our method, have $\Theta(p^3)$ time complexity.
Fig. \ref{fig:scalability} shows the scalability comparison of T-CorEx and these methods.

\textbf{Online Regime.} The temporal covariance estimation problem described above and the proposed algorithm both assume an offline regime -- i.e., the data of all time-steps are received at once.
Nevertheless, T-CorEx can be applied in an online fashion as well.
One straightforward way of doing this is to first set the sample weights $\alpha_t(\tau)$ to zero whenever $\tau > t$ (i.e., not allowing to use samples from the future) and then, as we receive data for a new time period, train the parameters of the linear CorEx of this new time period, possibly fine-tuning the parameters of the previous $h > 0$ time periods.

\section{Experiments}\label{sec:experiments}
We compare T-CorEx with other static and temporal covariance estimation methods on both synthetic (Sec.~\ref{subsec:syn-data}) and real-word datasets (Secs. ~\ref{subsec:stocks} and \ref{subsec:fmri}).
Static baselines are the Ledoit-Wolf (LW) method \citep{ledoit2004well}, factor analysis (FA), sparse PCA \citep{zou2006sparse,mairal2009online}, linear CorEx \citep{steeg2017low}, graphical lasso (GLASSO) \citep{friedman2008sparse}, and latent variable graphical lasso (LVGLASSO) \citep{chandrasekaran2010latent}. The temporal baselines are time-varying graphical lasso (TVGL) \citep{hallac2017network} and latent variable time-varying graphical lasso (LTGL) \citep{TomasiLTGL}. In the experiments with synthetic data, we add two additional baselines: T-CorEx-no-reg (T-CorEx with $\lambda = 0$) and T-CorEx-simple (T-CorEx with $\beta=10^{-9}$) to show the importance of enforcing temporal consistency and the importance of introducing sample weights respectively.

We use scikit-learn implementations of LW, FA, and Sparse PCA methods.
We use the original implementations of linear CorEx, TVGL, and LTGL.
For GLASSO, we tried the scikit-learn implementation, the QUIC method (\url{http://www.cs.utexas.edu/~sustik/QUIC/}), and TVGL with $\beta=0$. In our experiment, the TVGL implementation was always better. Therefore, we selected the latter implementation.
For LVGLASSO, we used the implementation available in the REGAIN repository (\url{https://github.com/fdtomasi/regain}), which also contains the original implementation of LTGL.

In all experiments we split the data into train, validation and test sets.
We use the validation set to select hyperparameters and we report the final scores on the test set.
The evaluation metric we use in our experiments is the negative log-likelihood of estimates averaged over time periods.
We justify our choice of the metric with the fact that all baselines besides the Ledoit-Wolf method assume the data has Gaussian distribution.
The details of implementations of baselines and grids of hyperparameter values of baselines are presented in the appendix (Sec.~\ref{sec:app-experimental-details}).

\begin{table*}
\begin{center}
\resizebox{\textwidth}{!}{%
\begin{tabular}{lcccccccccc}
\toprule
\multirow{2}{*}{Method} & \multicolumn{5}{c}{Sudden Change} & \multicolumn{5}{c}{Smooth Change} \\ 
& $s=8$ & $s=16$ & $s=32$ & $s=64$ & $s=128$ & $s=8$ & $s=16$ & $s=32$ & $s=64$ & $s=128$\\
\midrule
Ground Truth & 196.0 & 196.0 & 196.0 & 196.0 & 196.0 & 230.2 & 230.2 & 230.2 & 230.2 & 230.2 \\
LW         & 286.2 & 266.7 & 252.8 & 239.9 & 227.5 & 287.3 & 278.9 & 270.1 & 262.7 & 254.9 \\
FA         & - & 524.4 & 236.9 & 208.5 & 201.3 & - & 630.5 & 267.5 & 242.3 & 235.6 \\
Sparse PCA & 270.5 & 232.9 & 212.8 & 205.3 & 200.7 & 274.7 & 260.1 & 247.4 & 238.8 & 235.4 \\
Linear CorEx & 312.3 & 221.5 & \textbf{204.6} & \textbf{199.7} & \textbf{197.7} & 333.0 & 267.8 & 243.0 & 235.5 & \textbf{233.0} \\
GLASSO     & 266.5 & 242.0 & 221.3 & 212.3 & 205.5 & 280.1 & 262.5 & 249.2 & 241.9 & 238.3 \\
LVGLASSO   & 271.6 & 245.5 & 235.5 & 217.6 & 210.2 & 276.6 & 267.1 & 254.7 & 248.5 & 240.7 \\
TVGL       & 237.5 & 224.5 & 213.4 & 207.6 & 203.7 & 259.0 & 251.9 & 244.0 & 239.4 & 236.7 \\
LTGL       & 248.6 & 230.0 & 218.7 & 209.6 & 204.7 & 265.0 & 256.2 & 247.1 & 241.8 & 238.7 \\
T-CorEx    & \textbf{228.0} & \textbf{213.8} & 205.3 & \textbf{199.6} & \textbf{197.7} & \textbf{250.6} & \textbf{243.3} & \textbf{237.6} & \textbf{234.5} & \textbf{232.7} \\
\midrule
T-CorEx-simple & 275.8 & 217.9 & \textbf{204.7} & \textbf{199.7} & \textbf{197.7} & 294.5 & 261.3 & 242.5 & 235.8 & \textbf{233.1} \\
T-CorEx-no-reg & 245.3 & 228.7 & 207.5 & \textbf{199.6} & \textbf{197.8} & 259.2 & 252.5 & 241.6 & 235.4 & \textbf{232.9} \\
\bottomrule
\end{tabular}
}
\end{center}
\caption{Time-averaged negative log-likelihood of estimates on synthetic data with sudden/smooth change. The data of each time period is generated from a modular latent factor model with $m=8$ and $p=128$.}
\label{tab:syn_results}
\end{table*}

\subsection{Synthetic Data}\label{subsec:syn-data}
We design experiments with synthetic data to test our method in the case when the data of each period is generated from a modular model.
We generate synthetic data for two scenarios.
In the first scenario, the underlying covariance matrix is constant for the first half of the time periods, then a sudden change happens after which the underlying covariance matrix remains constant for the remaining half of time periods.
We call this scenario \textit{sudden change}.
In the second scenario, the underlying covariance matrix is slowly changing from $\Sigma_1$ to $\Sigma_2$.
We call this scenario \textit{smooth change}.

To describe a Gaussian modular latent factor model, it is enough to specify 6 quantities: $\mathbb{E}[Z_j], \mathbb{E}[Z^2_j]$, $\mathbb{E}[X_i]$, $\mathbb{E}[X^2_i]$, $\pi_i$ -- the parent of $X_i$, and $\rho_i$ -- the correlation of $X_i$ and its parent.
Note the moments of $Z_j$ do not affect the marginal distribution of $X$, so w.l.o.g. we set $\mathbb{E}[Z_j] = 0$ and $\mathbb{E}[Z^2_j]=1$.
We also set $\mathbb{E}[X_i] = 0$.
In our experiments we have that $\pi_i \sim \text{Uniform}\{1,2,\ldots,m\}$ and $\sigma_i \sim \text{Uniform}[1/4, 4]$.
To define the correlations $\rho_i$, for each $X_i$ we first sample the signal-to-noise ratio ($\text{snr}$) of $X_i \mid  Z_{\pi_i}$ uniformly from [0, 5] and then set the correlation between $X_i$ and $Z_{\pi_i}$, $\rho_i = \text{sgn}(\xi) \sqrt{\frac{\text{snr}}{\text{snr}+1}}$, with $\xi \sim \mathcal{N}(0,1)$. This way we control the average signal-to-noise ratio similar to the experiments done in \citep{steeg2017low}.

To create data with a sudden change, we generate two different modular models, $p^{(1)}(x,z)$ and $p^{(T)}(x,z)$.
The data of the first five periods is generated from $p^{(1)}(x,z)$, while the data of the next five periods is generated from $p^{(T)}(x,z)$.
We generate $s$ training, 16 validation and 1000 testing samples for each period.
The left multi-column of Table \ref{tab:syn_results} shows the comparison of baselines on this type of data for $m=8, p=128$ and varying values of $s$.

To create data with a smooth change, we generate two modular models, $p^{(1)}(x,z)$ and $p^{(T)}(x,z)$. 
Let the former be characterized by $\pi^{(1)}, \rho^{(1)}, \sigma^{(1)}$, and the latter be characterized by $\pi^{(T)}, \rho^{(T)}, \sigma^{(T)}$.
We start from $p^{(1)}(x,z)$ and smoothly change the model into $p^{(T)}(x,z)$, so that for each time period $t = 2, 3, \ldots, T-1$ the joint distribution remains modular. We define the parameters of $p^{(t)}(x,z)$ the following way: 
\begin{align*}
\rho^{(t)} &\triangleq \alpha_t \rho^{(1)} + (1-\alpha_t) \rho^{(T)}, \\
\sigma^{(t)} &\triangleq \alpha_t \sigma^{(1)} + (1-\alpha_t) \sigma^{(T)},
\end{align*}
with $\alpha_t = (T-t) / (T-1)$.
To define $\pi^{(t)}_i$, we randomly select when each variable will change its parent from $\pi^{(1)}_i$ to $\pi^{(T)}_i$. Formally, we sample $\tau_i \sim \text{Uniform}\{2,3,\ldots,T\}$ and set $\pi^{(t)}_i = \pi^{(1)}_i$ if $t < \tau_i$, otherwise we set $\pi^{(t)}_i = \pi^{(T)}_i$. 
We generate $s$ training, 16 validation and 1000 testing samples for each period from the corresponding $p^{(t)}(x,z)$.
The right multi-column of Table \ref{tab:syn_results} shows the comparison of baselines on this type of data for $m=8, p=128$ and varying values of $s$.

The results of sudden change and smooth change experiments show that the proposed method gives state-of-the-art results when data is generated from a modular model.
The advantage of T-CorEx is significant when the number of samples is small. 
We see that both T-CorEx-simple and T-CorEx-no-reg improve over linear CorEx, but are significantly worse than T-CorEx.
We conclude that both the temporal consistency regularization term and introducing sample weights are needed to get the best performance.

We also consider generating synthetic datasets $m=32$ latent factors (instead of 16) and $p=128$ observed variables.
This way we make the ground truth covariance matrices 4 times sparser.
Table~\ref{tab:syn_results_m32} presents the results for this case.
We see that T-CorEx still outperform the other baselines, but with smaller margin compared to the experiments with $m=16, p=128$.
This behavior is expected, as by increasing the sparsity level, we give preference to the TVGL and LTGL methods, which make sparsity assumptions.
In our experiments we also noticed that decreasing $m$ increases the gap between T-CorEx and other baselines, since the covariance matrices become less sparse.

\subsubsection{Clustering of Observed Variables}
\begin{figure}[th]
    \centering
    \includegraphics[width=0.65\textwidth]{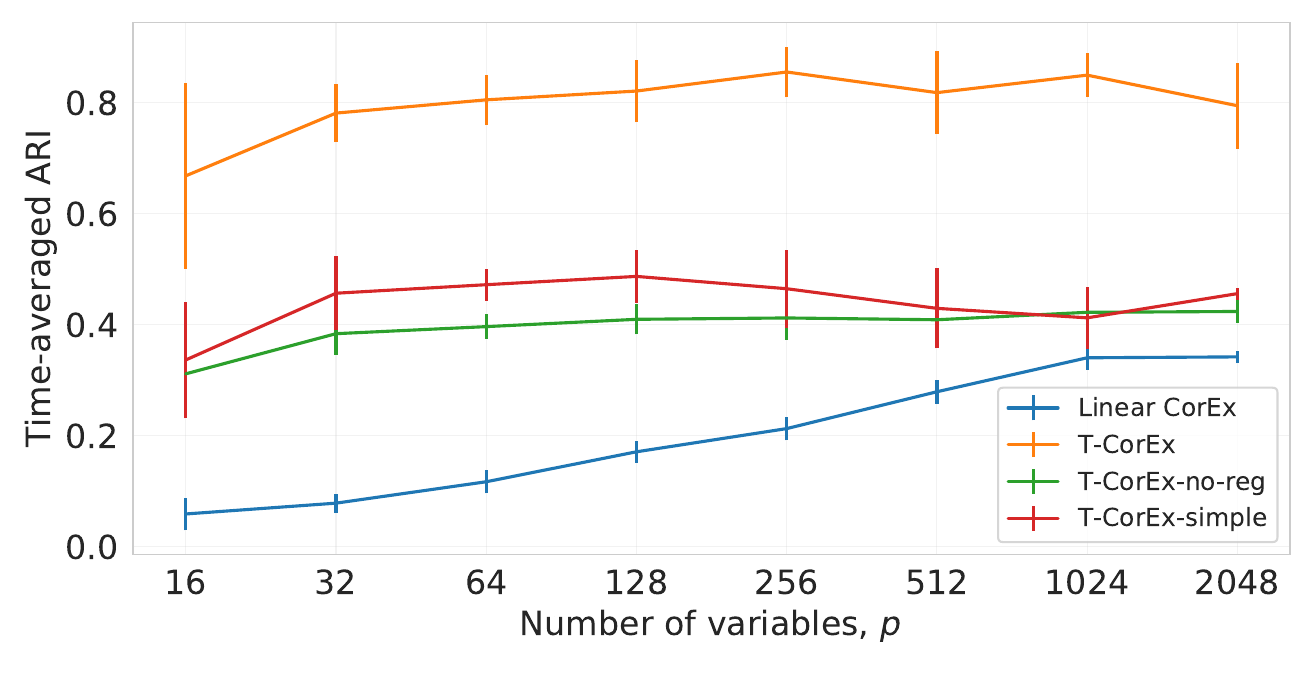}
    \caption{Evidence of blessing of dimensionality effect when learning modular latent factor models. We report the time-averaged adjusted Rand index. Error bars are standard deviation over 10 runs.}
    \label{fig:tcorex-blessing}
\end{figure}

Beside covariance estimation, T-CorEx can also be used for temporal clustering of observed variables. 
This is done by grouping variables by the hidden variable that has highest mutual information with it.
As it will be shown in Sec.~\ref{subsec:fmri}, this kind of analysis helps to group variables that vary in a correlated way.

It has been shown that Linear CorEx exhibits blessing of dimensionality when the data comes from a modular or an approximately modular latent factor model~\citep{steeg2017low} -- for a fixed number of latent factors the clustering performance of Linear CorEx increases as the number of observed variables increases.
In this section we test whether T-CorEx also exhibits blessing of dimensionality.
We consider sudden-change synthetic data (see Sec.~\ref{subsec:syn-data}), with $T=10$ time periods, $m=8$ latent factors, and 8 training samples per period.
We vary the number of observed variables, $p$, and measure the quality of clusters using the time-averaged adjusted Rand index (ARI), which is adjusted for chance to give 0 for a random clustering and 1 for a perfect clustering. 
The results presented in Fig.~\ref{fig:tcorex-blessing} tell multiple things at once.
First, T-CorEx performs significantly better than its simpler variants: T-CorEx-simple and T-CorEx-no-reg.
This confirms our conclusion that both temporal consistency regularization and sample weighting are essential for T-CorEx.
Second, when $p\le 256$ T-CorEx also exhibits blessing of dimensionality.
The performance starts to plateau when $p\ge 512$, which might be due to suboptimal optimization.

\subsection{Stock Market Data}\label{subsec:stocks}
\begin{figure*}[t]
    \centering
    \includegraphics[width=\textwidth]{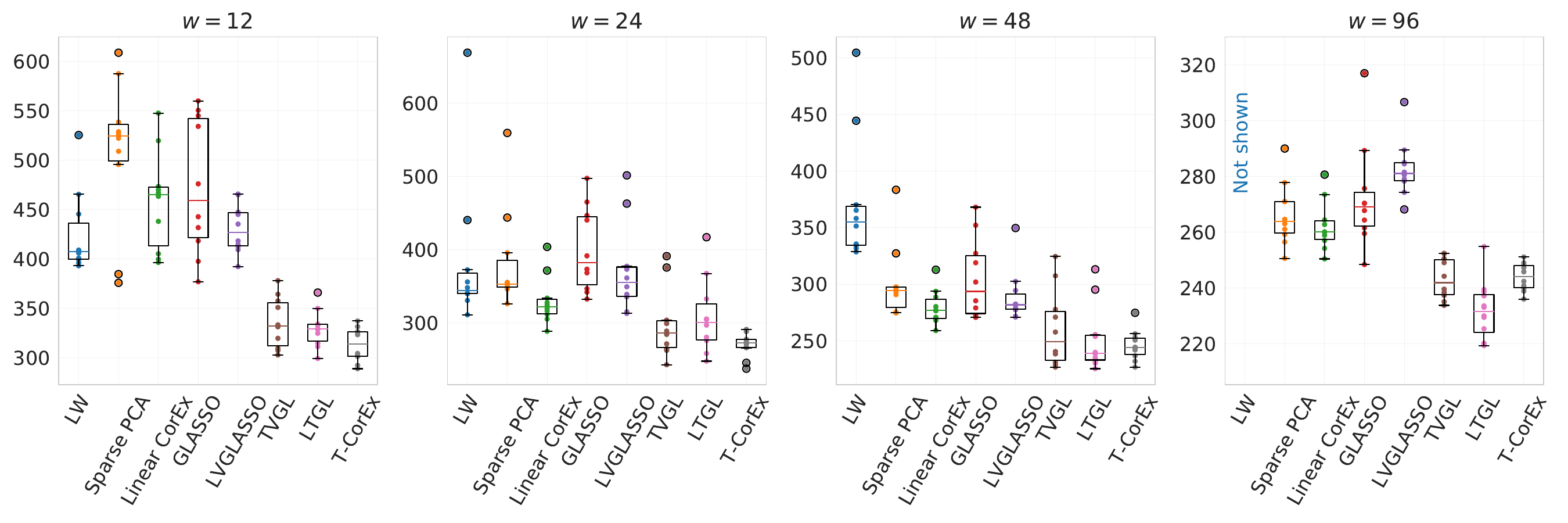}
    \caption{Time-averaged negative log-likelihood of estimates on stock market test data for 10 random train/val/test splits. The factor analysis baseline is not shown to keep the plots readable.}
    \label{fig:10-runs}
\end{figure*}
\begin{figure*}[!h]
    \centering
    \begin{subfigure}[t]{0.32\textwidth}
        \centering
        \includegraphics[width=\textwidth]{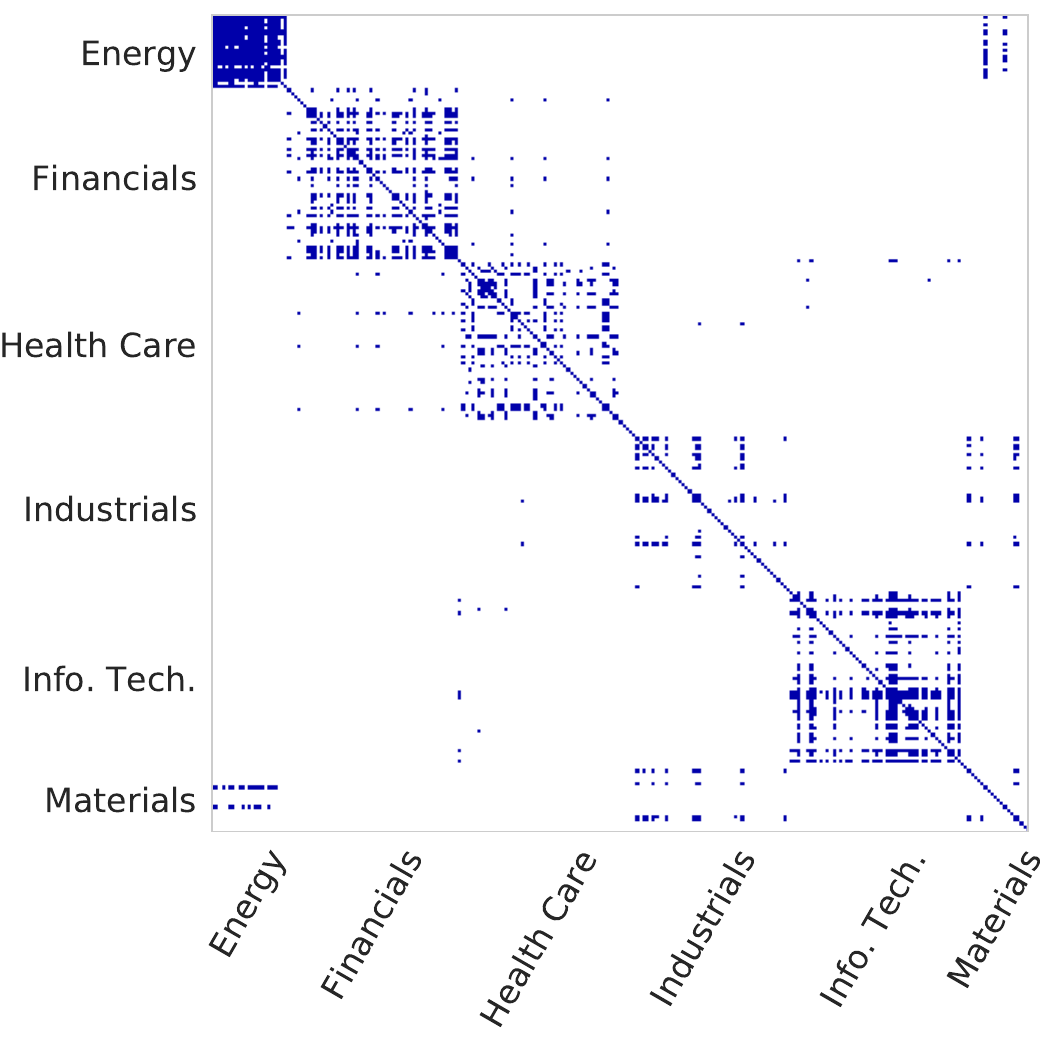}
        \label{fig:lc-whole}
    \end{subfigure}%
    ~ 
    \begin{subfigure}[t]{0.32\textwidth}
        \centering
        \includegraphics[width=\textwidth]{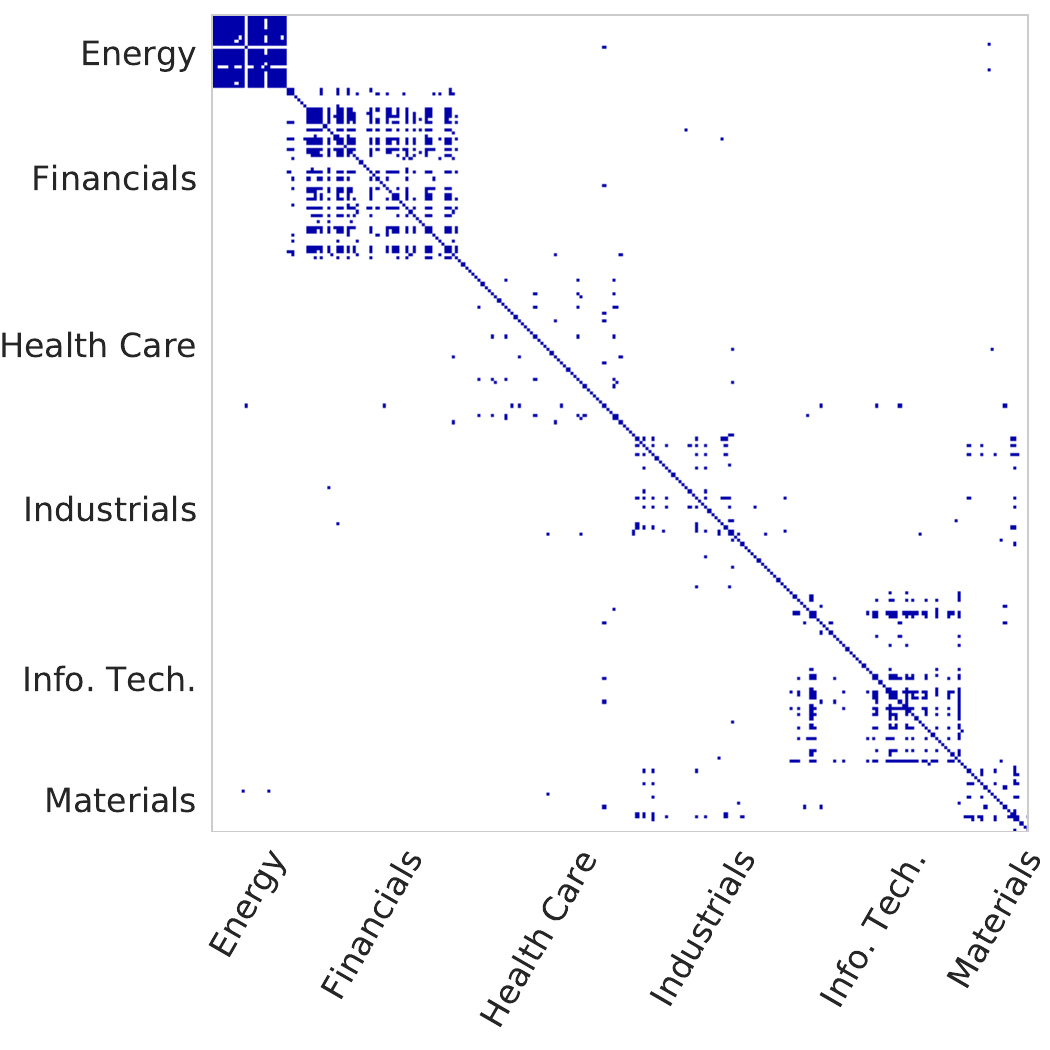}
        \label{fig:tc-radnom-point}
    \end{subfigure}%
    ~
    \begin{subfigure}[t]{0.32\textwidth}
        \centering
        \includegraphics[width=\textwidth]{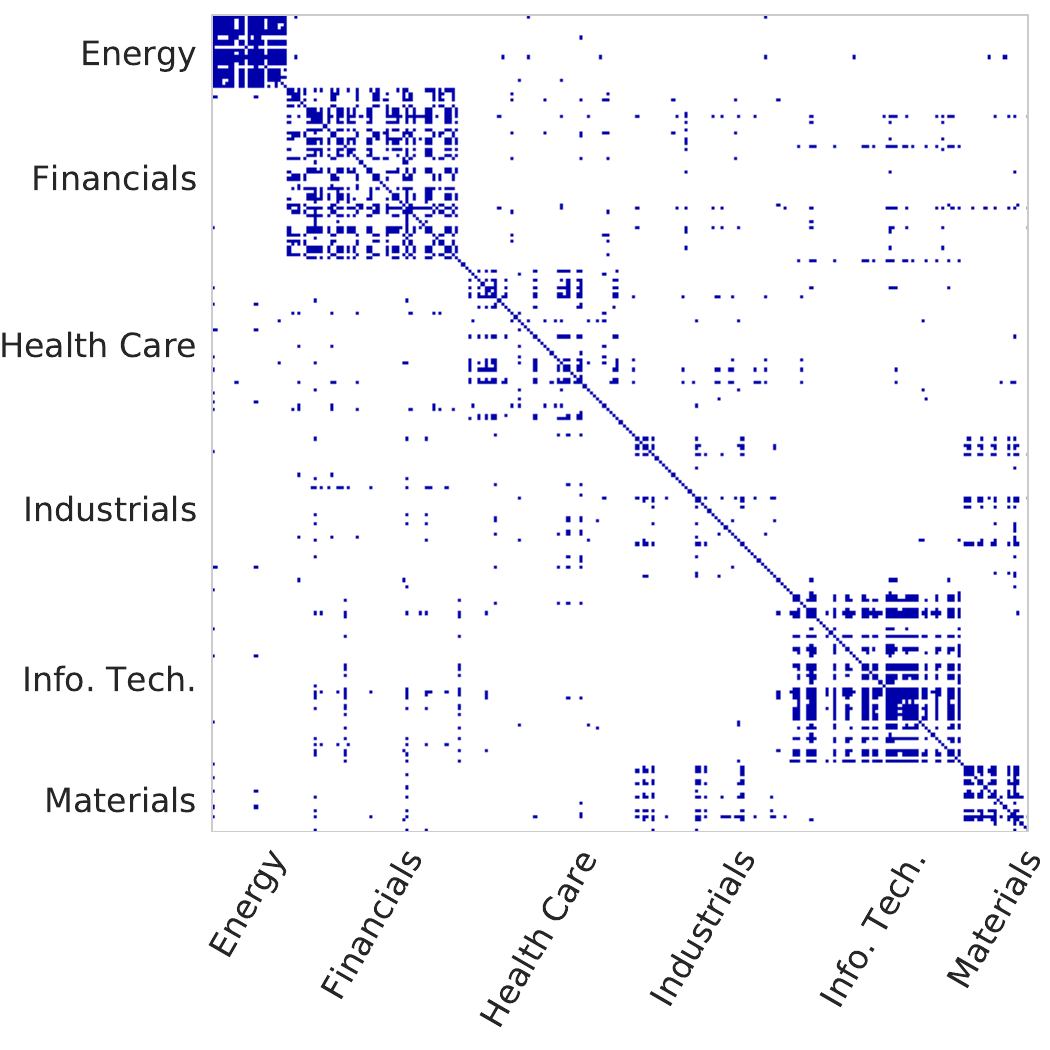}
        \label{fig:lc-random-point}
    \end{subfigure}
    \caption{Thresholded inverse correlation matrices on the stock market data for three cases. \textbf{Left:} the estimate of linear CorEx trained on whole data (2000-2016). \textbf{Middle:} the estimate of T-CorEx for a random two-week period. \textbf{Right:} the estimate of linear CorEx trained on 128 samples around the same two-week period.
    For visualization purposes not all sectors are shown.}
    \label{fig:inv_covs}
\end{figure*}
Next, we evaluate our method on a stock prices dataset.
We take the daily sampled historical prices of S\&P 500 stocks from January, 2000 to January, 2016.
For simplicity, we keep the stocks that are present for at least 95\% of the period. 
The resulting dataset contains prices of 391 stocks. 
We compute the daily log-returns, standardize each variable, and add isotropic Gaussian noise with $10^{-4}$ variance.
We consider four period sizes: $w\in\{12, 24, 48, 96\}$.
For each period size we randomly partition samples of each period into train, validation and test sets containing $2w/3, w/6$, and $w/6$ samples respectively.
For computational efficiency of experiments, we consider only the last 10 periods.

Fig.~\ref{fig:10-runs} shows the results on 10 random train/validation/test splits for each period size (please refer to Table~\ref{tab:stock-day} of the appendix for the exact values).
T-CorEx is doing better than the other methods when the number of samples is small ($w=12, 24$).
For $w=48$, we see that TVGL, LTGL and T-CorEx perform similarly.
At $w=96$ LTGL starts to produce better estimates.
We hypothesize that the assumption of T-CorEx of data distribution being approximately modular is more restrictive than the sparsity assumption of LTGL, but it does not introduce large biases for stock market data.
This explains why T-CorEx gives better results when the period size is small and why T-CorEx has the smallest standard deviation of scores.


\subsubsection{Qualitative Analysis}\label{subsec:qualitative}
Next, we want to find what are the qualitative differences between estimates of T-CorEx and linear CorEx on the stock market data.
For this purpose we plot the entries of inverse correlation matrix that have absolute value greater than some threshold, which is set to 0.025 in our experiments.
This can be interpreted as plotting the adjacency matrix of a Markov random field. 
First, we train a linear CorEx on the whole period, ignoring the temporal aspect (the left part of Fig. \ref{fig:inv_covs}). This shows how the system looks like ``on average.''
We see that most of the edges are within sectors. Then we train T-CorEx with period length equal to two weeks and plot its estimate for a random time period $t_0$ (the middle part of Fig. \ref{fig:inv_covs}).
First, we see that the sectors are not as densely connected.
Second, T-CorEx captures some dependencies that are not present for the whole period.
The opposite is also true---two variables can be directly connected on average, but be disconnected for some period of time (e.g., some connections between sectors ``Materials'' and ``Energy'').
The advantage of T-CorEx lies in the ability of detecting correlations that occur only for a short period of time. Methods requiring large number of samples (large period size $w$) cannot detect such correlations.
To finalize our analysis, we fit linear CorEx on $s$ samples around the same random time period $t_0$.
When $s$ is too small, the estimates are too noisy.
When $s$ is too large the estimates are less related to the true statistics of time period $t_0$.
The right part of Fig. \ref{fig:inv_covs} shows the estimate of linear CorEx for $s=128$ (the best value according to log-likelihood).
The estimate of this linear CorEx is qualitatively different from that of T-CorEx and gives a significantly worse score, indicating the inherent problems of applying a static covariance estimation independently to each time period.

\begin{figure*}[!t]
    \centering
    \includegraphics[width=\textwidth]{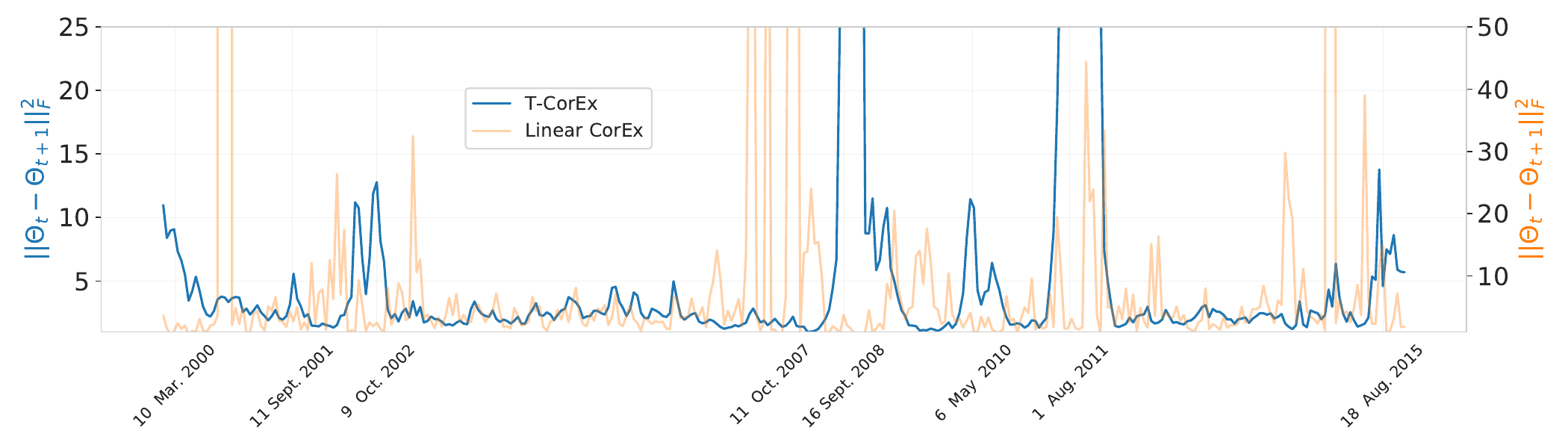}
    \caption[Caption for LOF]{Frobenius norm of the difference of inverse correlation matrices $\Theta_t$ of neighboring time periods for T-CorEx (left axis) and linear CorEx (right axis). The marked events on $x$ axis correspond to major US stock market events (\url{en.wikipedia.org/wiki/List_of_stock_market_crashes_and_bear_markets}) --- the collapse of a technology bubble (10 March 2000), the stock market downturn of 2002 (9 Oct 2002), the start of the US bear market of 2007-09 (11 Oct 2007), the financial crisis of 2007-08 (16 Sep 2008), the 2010 flash crash (6 May 2010), the August 2011 stock markets fall (1 Aug 2011), and the 2015-16 stock market selloff (18 Aug 2015).}
    \label{fig:chane-point-detection}
\end{figure*}

\subsubsection{Change Point Detection}
Estimated covariance matrices can be used to detect large changes in the underlying structure.
One simple way to do this is to look at the Frobenius norm of the difference of inverse correlation matrices of neighboring time periods.
Fig. \ref{fig:chane-point-detection} shows that T-CorEx detects all but one of the major US stock market events happened between 2000 and 2016.
For some events major changes are visible up to two months earlier.
Fig. \ref{fig:chane-point-detection} also shows the same analysis for linear CorEx with optimal period size.
The peaks of linear CorEx align much worse with the ground truth events.
Furthermore, the estimates of linear CorEx are temporally less consistent.
As expected, linear CorEx does not detect changes that occur in very short periods, such as the changes related to the September 11 attacks and the 2010 flash crash.

\subsection{High-Resolution fMRI Data}\label{subsec:fmri}
\begin{figure}
    \centering
    \includegraphics[width=0.65\textwidth]{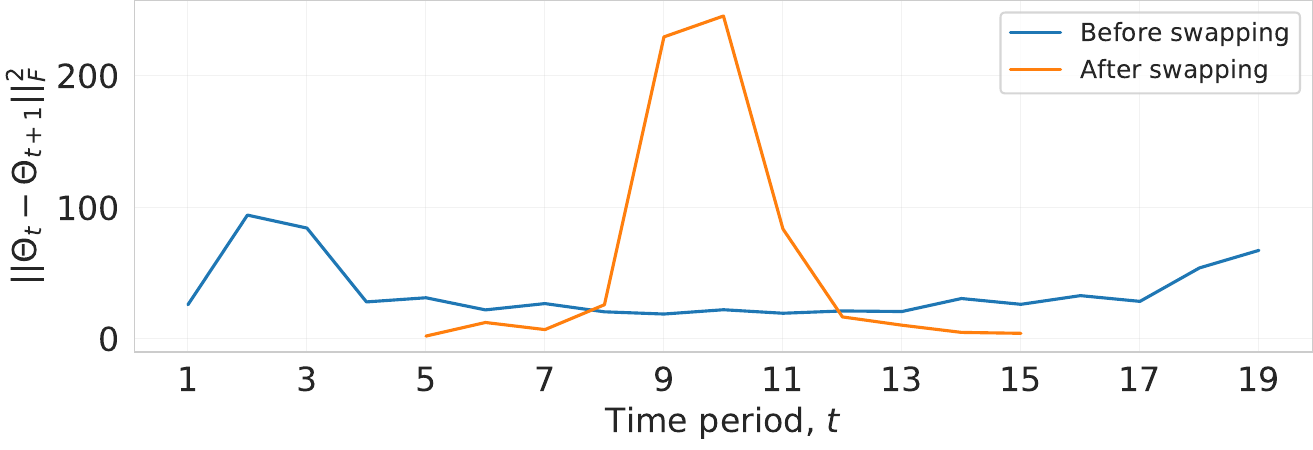}
    \caption{\textbf{Blue:} Frobenius norm of the difference of inverse correlation matrices $\Theta_t$ of neighboring time periods for a resting state fMRI session. \textbf{Orange:} the same statistics for a T-CorEx trained on the middle 12 time periods after swapping its first 6 and last 6 periods.}
    \label{fig:fmri}
\end{figure}

\begin{figure*}
    \centering
    \small
    \begin{subfigure}{0.3\textwidth}
        \centering
        \includegraphics[width=\textwidth]{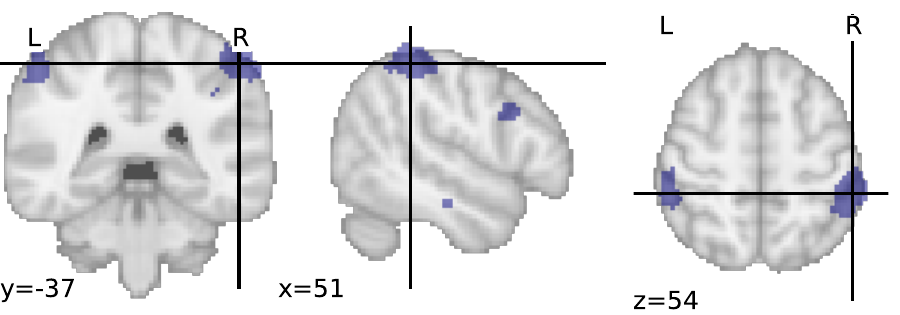}
        \caption{Right Brodmann area 40, with symmetric Left Brodmann area 40 activation.}
    \end{subfigure}%
    ~
    \hspace{0.5cm}
    \begin{subfigure}{0.3\textwidth}
        \centering
        \includegraphics[width=\textwidth]{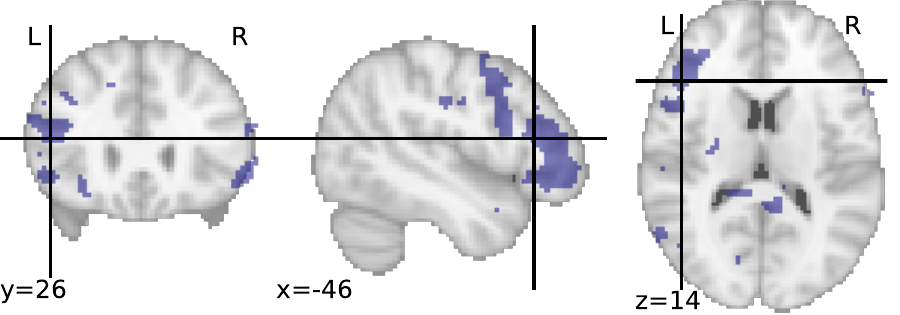}
        \caption{Left Brodmann area 44/45 (Broca's area), with asymmetric activations.}
    \end{subfigure}%
    ~
    \hspace{0.5cm}
    \begin{subfigure}{0.3\textwidth}
        \centering
        \includegraphics[width=\textwidth]{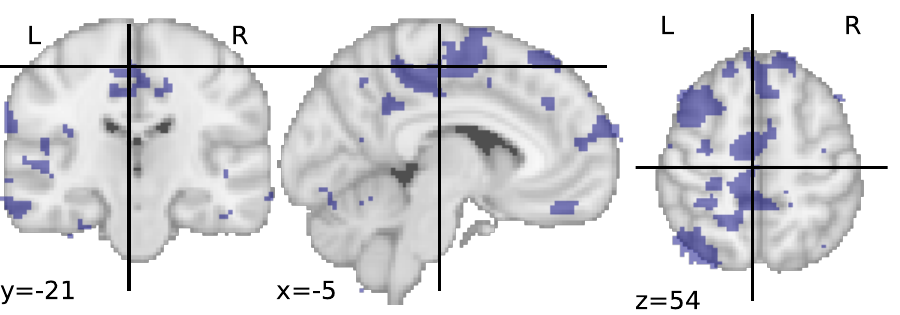}
        \caption{Left Brodmann area 6, with asymmetric frontal and parietal activations.}
    \end{subfigure}
    \normalsize
    \caption{Some of the clusters found by T-CorEx for time period 12. 
    }
    \label{fig:fmri-clusters}
\end{figure*}
To demonstrate the scalability and usefulness of T-CorEx, we apply it to high-dimensional functional magnetic resonance images (fMRI) data.
The standard measurement in fMRI is blood oxygen level-dependent contrast, which measures blood flow changes in biological tissues (``activation'') and is used to find active regions.
Usually, an fMRI session lasts 3-10 minutes, during which a few hundred high-resolution brain images are captured, each having 100K+ volumetric pixels (voxels).
Correlation analysis is widely used to study functional connections between brain regions~\citep{preti2017dynamic}.
While in general these analyses are conducted assuming static covariance, recently time-varying covariance (``dynamic functional connectivity'') has received more attention~\citep{chang2010time, rfMRI2018}.
This latter case is exactly the use case of T-CorEx.

We demonstrate the feasibility of T-CorEx in fMRI analysis by inducing an artificial change point in an otherwise stable time series.
While the induced shift is clearly synthetic, this experiment shows possible value for the fMRI community in detecting natural change points and/or using T-CorEx for more nuanced domain-specific analyses, demonstrating that T-CorEx can scale to the 100K+ variable regime.
First, we fit T-CorEx on the pre-processed resting-state fMRI session 014 of the MyConnectome project \citep{poldrack2015long}.
We then do spatial smoothing using a Gaussian filter with fwhm=8mm.
The session has 518 time-point volumes, each having 148262 voxels. We divide the whole period into 20 non-overlapping periods, ignoring the first 18 time-points.

The blue curve in Fig. \ref{fig:fmri} shows the Frobenius norm of differences of inverse correlation matrices of neighboring time periods.
Note that, although the correlation matrices are extremely large, we are able to compute the inverses and norms of differences (Sec.~\ref{sec:app-fast-operations}).
We see relatively large variability in the beginning and in the end of the session.
Next, we consider the middle 12 periods (i.e., removing 4 periods from both sides). We swap the first 6 and the last 6 periods of those periods, creating an artificial change in the middle, and retrain T-CorEx on the resulting data. 
The orange plot of Fig. \ref{fig:fmri} shows the Frobenius norm of differences of inverse correlation matrices of neighboring time periods for this case.
T-CorEx detects the shift we created, while other methods do not scale to this regime.

T-CorEx can also provide secondary analyses by grouping correlated regions.
We assign each voxel to the latent factor that has the largest mutual information with it, forming groups by each factor.
This produces $m=50$ clusters of voxels for each time period.
Fig. \ref{fig:fmri-clusters} shows three clusters from time period 12, identified using NeuroSynth~\citep{neurosynth}.
Clusters displayed in (a) and (b) correspond to standard anatomic regions \citep{brodmann1909vergleichende}, namely both left and right Brodmann area 40, as well as left Brodmann areas 44/45 (Broca's area), which is known to be a highly lateralized (asymmetric) region. Cluster (c) displays distributed activation in the left hemisphere. 
We found that cluster (a) is present in all time periods; (b) is present starting at time period 3. These two clusters exhibit small variations over time. Cluster (c) is present starting at time period 11, but varies more compared to the other two clusters.

\subsection{Hyperparameter Sensitivity Analysis of T-CorEx}\label{sec:app-tcorex-hyperparameters}
\begin{figure}[!t]
    \centering
    \includegraphics[width=0.98\columnwidth]{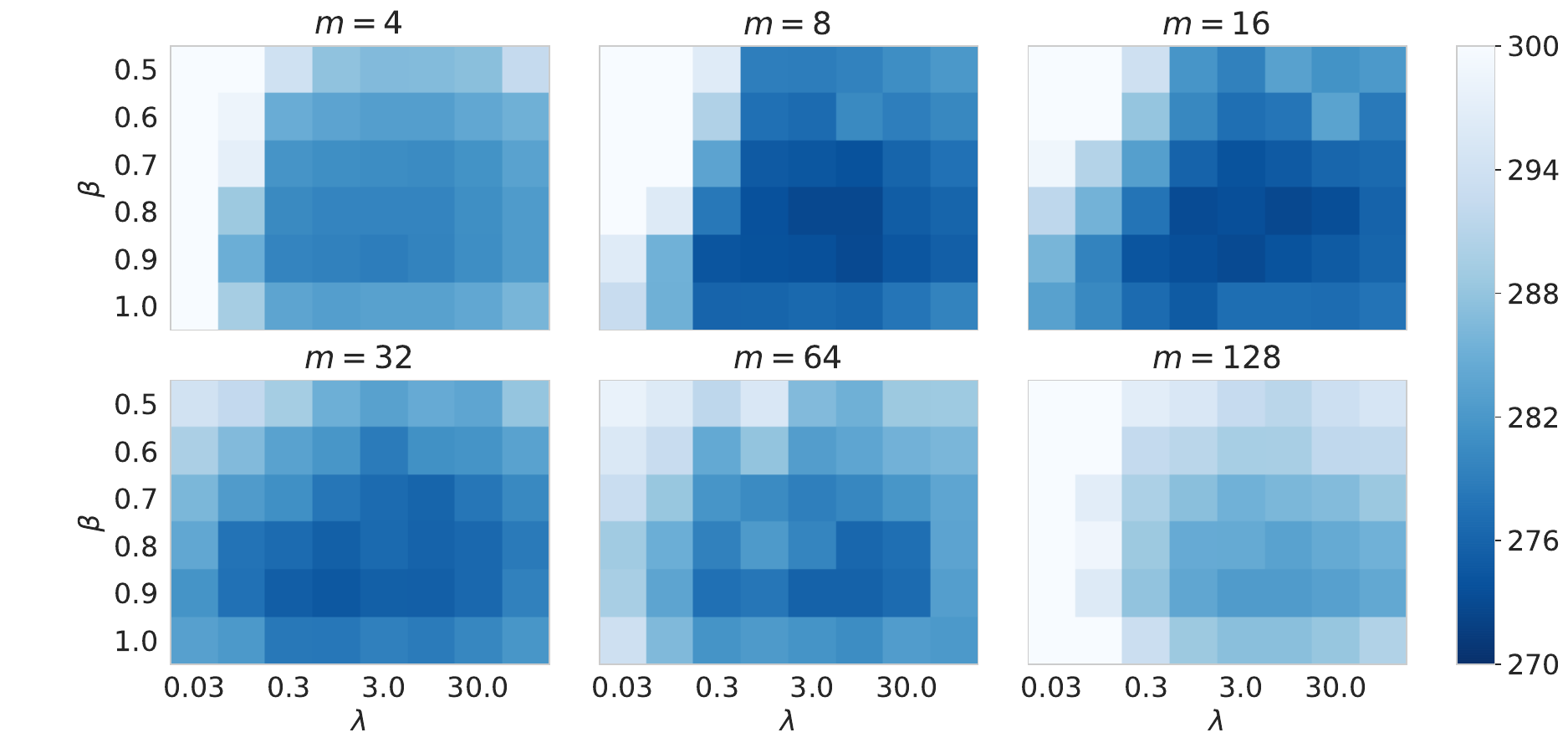}
    \caption{Performance of T-CorEx with various hyperparameter settings on stock market data with period size $w=24$ (Sec.~\ref{subsec:stocks}).
    The reported score is the time-averaged negative log-likelihood of estimates on the test data. The $\lambda$ axis is in logarithmic scale.}
    \label{fig:tcorex-hypers}
\end{figure}
In this subsection we analyze the hyperparameter sensitivity of T-CorEx and provide suggestions on setting hyperparameters.
T-CorEx has 4 hyperparameters: the number of latent variables $m$, the sample importance decay rate $\beta$, the regularization coefficient $\lambda$, and the regularization function $\Phi$.
The latter can be selected using our prior knowledge of the nature of changes the variables exhibit.
The $\ell_2$ regularization is better suited for smoothly varying time series, while the $\ell_1$ regularization is better for time series with sudden changes.

Our prior knowledge can also help to select $m$.
For example, in stock market data we know that there are 11 large sectors.
We can assume that the best value of $m$ is between 10 and 30.
Furthermore, setting $m$ to larger values than the best value should only help us to improve the T-CorEx objective, since the method can simply ignore the additional latent variables by setting the associated weights to zero.
However, it practice, overly large values require more training iterations and more computation.

The remaining two hyperparameters: $\beta$ and $\lambda$, have similar objectives, as both try to enforce temporal consistency of learned models.
In all our experiments we observed that the best value of $\lambda$ lies in between $0.1$ and $10.0$.
Furthermore, overly large values of $\lambda$ did not degrade the performance too much.
Also, we observed that the best value of $\beta$ depended on the period size significantly.
For small period sizes $(w \le 16)$ large values of $\beta$ worked better ($\beta\in [0.8,1.0]$, whereas for relatively large values of $w$ ($w \ge 64$) we observed that small values of $\beta$ work better ($\beta \in [0.3, 0.5]$).
To understand the interplay between $\beta, \lambda$ and $m$ we present the performance of T-CorEx with various hyperparameter settings in Fig.~\ref{fig:tcorex-hypers}.
The data is the stock market data with period size $w=24$ described in Sec.~\ref{subsec:stocks}.
First of all, we see that the best values of $\beta$ and $\lambda$ do not change much depending on $m$.
This simplifies model selection as one can choose $m$ independently from  $\beta$ and $\lambda$.
Furthermore, as expected, overly large values of $m$ degrade the performance.
More importantly, Fig.~\ref{fig:tcorex-hypers} shows that when $m$ is selected well ($m=8,16$) the performance is less sensitive to the values of $\beta$ and $\lambda$.

\section{Related Work}\label{sec:related}
Many works have addressed the problem of high-dimensional covariance estimation \citep{fan2016overview}.
One major direction of estimating high-dimensional covariance matrices is introducing sparsity constraints.
Sparse covariance matrix estimation for multivariate Gaussian random variables is investigated in \citep{tibshirani_sparse_cov}.
Most often sparsity constraints are imposed on the inverse covariance matrix (precision matrix).
The precision matrix encodes the conditional independences between pairs of variables.
Learning a sparse precision matrix corresponds to learning a sparse graphical model~\citep{banerjee2008model}.
The graphical lasso method~\citep{friedman2008sparse} does sparse inverse covariance matrix estimation for multivariate Gaussian random variables.
The problem of network inference using graphical lasso is well studied~\citep{meinshausen2006high,yuan2007model, friedman2008sparse, ravikumar2011high}. 

In many real-world applications, there are latent factors influencing the system, modeling which usually leads to better estimators.
Factor analysis and probabilistic PCA \citep{tipping1999probabilistic} are latent factor models that can be used for covariance estimation.
Unfortunately, they fail in undersampled regimes.
Sparse PCA \citep{zou2006sparse,mairal2009online} remedies this problem.
Covariance estimation can also be done by learning graphical models with latent factors \citep{choi2010gaussian,chandrasekaran2010latent,choi2011learning}.
The latent variable graphical lasso method \citep{chandrasekaran2010latent} learns a sparse graphical model with latent factors.
Linear CorEx \citep{steeg2017low} is another latent factor model and is central to this work.

Many works extended a particular covariance estimation method for temporal covariance estimation.
Sparse PCA has been adapted for high-dimensional multivariate vector autoregressive time series \citep{wang2013sparse}.
The time-varying graphical lasso method (TVGL) \citep{hallac2017network} extends graphical lasso.
The latent variable time-varying graphical lasso (LTGL) \citep{TomasiLTGL} extends latent variable graphical lasso.
While T-CorEx extends linear CorEx in a similar fashion, there are key differences.
Instead of encouraging sparse solutions, T-CorEx encourages approximately modular models.
Additionally, T-CorEx uses weighted samples of other time periods.
Most importantly, T-CorEx scales linearly with the number of observed variables $p$, while TVGL and LTGL scale cubically and store $p\times p$ matrices.
This makes them inapplicable for systems having more than $10^4$ variables.
To our knowledge, T-CorEx is the only temporal covariance estimation method that has linear time and memory complexity with respect to $p$.

\section{Conclusion}\label{sec:conclusion}
We developed a novel temporal covariance estimation method called T-CorEx.
The proposed method has linear time stepwise computational complexity with respect to the number of observed variables assuming $m$ does not depend on $p$.
We evaluated our method on both synthetic and real-world datasets, showing state-of-the-art results in highly undersampled regimes.
We also studied the range of possible applications of T-CorEx.
In future research we aim to simplify the hyperparameter selection of T-CorEx.
One way to achieve this is to learn the sample weights $\alpha_t(\tau)$ automatically.

\section*{Acknowledgments}
The authors would like to thank Federico Tomasi for his valuable help on training latent variable time-varying graphical lasso, and Neal Lawton for insightful discussions.
This research is based upon work supported in part by DARPA, via W911NF-16-1-0575, and the Office of the Director of National Intelligence (ODNI), Intelligence Advanced Research Projects Activity (IARPA), via 2016-16041100004. The views and conclusions contained herein are those of the authors and should not be interpreted as necessarily representing the official policies, either expressed or implied, of DARPA, ODNI, IARPA, or the U.S. Government. The U.S. Government is authorized to reproduce and distribute reprints for governmental purposes notwithstanding any copyright annotation therein. Additionally, H. Harutyunyan was supported by the USC Annenberg Fellowship.

\begin{small}
\setlength{\bibsep}{0.25em}
\bibliography{main}
\bibliographystyle{apalike}
\end{small}

\appendices
\section{Efficient operations with diagonal plus low-rank matrices}\label{sec:app-fast-operations}
As discussed in the main text, covariance matrix estimates of T-CorEx are diagonal plus low-rank matrices of form $D + U^TU$, where $D \in \mathbb{R}^{p\times p}$ is a diagonal matrix and $U \in \mathbb{R}^{m \times p}$.
Throughout the paper we avoid constructing such covariance estimates explicitly as it requires $O(mp^2)$ time and $O(p^2)$ memory.
In this section, we describe how to do some important operations on such matrices in complexity that is linear in $p$.
First of all, multiplying such a matrix with a matrix of size $p \times k$ can be done in $O(m k p)$ time.
The determinant of such a matrix can be computed in $O(m^3 + m^2p)$ time using the generalization of the matrix determinant lemma:
\begin{equation*}
    \text{det}(D + U^TU) = \text{det}(I_m + UD^{-1}U^T)\text{det}(D).
\end{equation*}
Inverting a diagonal plus low-rank matrix $A=D+U^TU$ can be done in $O(m^3 + m^2p)$ using the Woodbury matrix identity. 
The inverse is $A^{-1} = D^{-1} - D^{-1} U^T (I_m + U D^{-1} U^T)^{-1} U D^{-1}$.
More importantly, this inverse can be written as a diagonal minus low-rank matrix of form $D_\text{inv} - V^T V$, with  $D_\text{inv} = D^{-1} \text{ and } V = R U D^{-1}$, where $R$ is an $m \times m$ matrix such that $R^T R = (I_m + U D^{-1} U^T)^{-1}$.
To find such $R$, we can use Cholesky decomposition of $(I_m + A^T D^{-1} A)^{-1}$, which can be done in $O(m^3)$ time.

\textbf{Frobenius norm of differences.} For change point detection experiments described in the main text, we needed to compute the Frobenius norm of the difference of two inverse correlation matrices.
Since correlation/covariance matrices produced by T-CorEx are diagonal plus low-rank matrices, their inverse can be written as diagonal minus low-rank matrices.
Let us consider two such matrices, $D_1 - U^TU$ and $D_2 - V^TV$.
Our task is to compute $\lVert (D_1 - U^TU) - (D_2 - V^TV) \rVert_F^2$ in linear time with respect to $p$.
We start with
\begin{align*}
\lVert &D_1 - U^TU - D_2 + V^TV \rVert_F^2 \\
&=\left\lVert\left((D_1 - D_2) + (V^TV - U^TU)\right) \right\rVert_F^2\\
&= \text{tr}\left(\left(D_1 - D_2\right) \left(D_1 - D_2\right)^T \right) \\
&\quad+2 \text{tr}\left(\left(D_1 - D_2\right)\left(V^TV - U^TU\right)^T\right) + \\
&\quad+\text{tr}\left(\left(V^TV - U^TU\right)\left(V^TV - U^TU\right)^T\right)\\
&= \lVert D_1 - D_2 \rVert_F^2 + 2 \text{tr}\left(\left(D_1 - D_2\right)\left(V^TV - U^TU\right)^T\right) \\
&\quad+ \lVert V^TV - U^TU \rVert^F_2.
\end{align*}
The first term is easy to compute in $O(p)$ time.
The middle term can be computed can be computed in $O(m p)$ time, because the diagonal terms of the matrix inside the trace operation are $(D_1 - D_2)_{i,i} (\langle v_i, v_i \rangle - \langle u_i, u_i \rangle)$, where $u_i$ and $v_i$ are the columns of $U$ and $V$ correspondingly.
Computing and summing these diagonal terms can be done in $O(m p)$ time.
It remains to compute the third term:
\begin{align*}
\lVert V^TV - U^TU \rVert^F_2 &= \text{tr}\left( \left(V^TV - U^TU\right) \left(V^TV - U^TU\right)^T \right)\\
&\hspace{-2cm}= \text{tr}\left(V^TVV^TV\right) - 2 \text{tr}\left(V^TVU^TU\right) + \text{tr}\left(U^TUU^TU\right)\\
&\hspace{-2cm}= \lVert VV^T \rVert_F^2 - 2 \lVert VU^T \rVert_F^2 + \lVert UU^T \rVert.
\end{align*}
All these 3 terms above can be computed in $O(m^2 p)$ time by constructing $UU^T$, $VU^T$ and $VV^T$ explicitly.

\textbf{Finding variables with largest changes.} Besides computing the overall change between two inverse correlation matrices, it is also interesting to find variables exhibit largest changes.
Given two diagonal minus low-rank matrices, $D_1 - U^TU$ and $D_2 - V^TV$, we need to compute $\sum_{i'=1}^p (D_1 - U^TU - D_2 + V^TV)_{i,i'}^2$ for each $i=1,\ldots,p$.
Again, let $u_i$ and $v_i$ denote the columns of $U$ and $V$ matrices correspondingly.
For fixed $i$, we have:
\begin{align*}
\sum_{i'=1}^p &(D_1 - U^TU - D_2 + V^TV)_{i,i'}^2 =\sum_{i'=1}^p\bigg( (D_1 - D_2)_{i,i'}^2 +. \\
&\hspace{-0.4cm}+ 2 (D_1 - D_2)_{i,i'} (V^TV - U^TU)_{i,i'} + (V^TV - U^TU)^2_{i,i'}\bigg).
\end{align*}
The first and second terms are easy to compute, because in each of them only one value is non-zero.
The third term can be written the following way:
\begin{align*}
\sum_{i'=1}^p (V^TV - U^TU)^2_{i,i'} &= \sum_{i'=1}^p\left( \langle u_i, u_{i'} \rangle^2 - 2 \langle u_i, u_{i'} \rangle \langle v_i, v_{i'} \rangle  +\right.\\
&\hspace{1cm}+\left.\langle v_i, v_{i'} \rangle^2\right)\\
&\hspace{-1.5cm}= \langle u^T_i U, u^T_i U \rangle - 2 \langle u^T_i U, v^T_i V \rangle + \langle v^T_i V, v^T_i V \rangle\\
&\hspace{-1.5cm}= u^T_i U U^T u_i - 2 u^T_i U V^T v_i + v^T_i V V^T v_i.
\end{align*}
Matrices $UU^T, VV^T$, and $UV^T$ are of size $m\times m$ and can be computed once in $O(m^2p)$ time.
Concluding we can computed the change of all variables in $O(m^2 p)$ time.

\begin{table*}[ht]
\begin{center}
\resizebox{\textwidth}{!}{%
\begin{tabular}{lcccccccccc}
\toprule
\multirow{2}{*}{Method} & \multicolumn{5}{c}{Sudden Change} & \multicolumn{5}{c}{Smooth Change} \\ 
& $s=8$ & $s=16$ & $s=32$ & $s=64$ & $s=128$ & $s=8$ & $s=16$ & $s=32$ & $s=64$ & $s=128$\\
\midrule
Ground Truth & 218.1 & 218.1 & 218.1 & 218.1 & 218.1 & 243.4 & 243.4 & 243.4 & 243.4 & 243.4  \\
LW         & 297.6 & 288.1 & 280.7 & 270.7 & 258.1 & 292.2 & 286.2 & 281.4 & 276.5 & 270.2  \\
FA         & - & - & - & 346.2 & 245.6 & - & - & - & 363.0 & 270.4  \\
Sparse PCA & 275.9 & 259.7 & 250.9 & 240.1 & 231.4 & 282.5 & 267.9 & 263.2 & 257.1 & 253.1  \\
Linear CorEx & 295.4 & 254.5 & 239.9 & 228.2 & \textbf{222.5} & 309.9 & 273.2 & 264.3 & 255.7 & 250.3  \\
GLASSO     & 281.2 & 256.9 & 244.0 & 235.8 & 228.3 & 287.5 & 267.4 & 261.0 & 253.8 & 251.5  \\
LVGLASSO   & 280.2 & 270.9 & 255.0 & 243.5 & 234.1 & 278.7 & 277.9 & 265.5 & 261.3 & 253.9  \\
TVGL       & 259.2 & \textbf{246.1} & 236.1 & 230.6 & 225.9 & 266.6 & 262.5 & 256.2 & 252.9 & 249.2  \\
LTGL       & 267.4 & 254.2 & 242.4 & 233.6 & 227.2 & 271.1 & 266.2 & 259.5 & 255.2 & 251.1  \\
T-CorEx    & \textbf{256.7} & 248.5 & \textbf{235.1} & \textbf{227.8} & 222.9 & \textbf{263.9} & \textbf{260.5} & \textbf{255.2} & \textbf{251.1} & \textbf{248.3}  \\
\midrule
T-CorEx-simple & 278.4 & 256.7 & 237.6 & \textbf{227.6} & \textbf{222.3} & 286.6 & 271.5 & 260.9 & 254.1 & 249.7  \\
T-CorEx-no-reg & 259.3 & 255.5 & 240.9 & 228.6 & 222.8 & 265.7 & 263.2 & 259.1 & 254.0 & 249.6  \\
\bottomrule
\end{tabular}
}
\end{center}
\caption{Time-averaged negative log-likelihood of estimates on synthetic data with sudden/smooth change. The data of each time period is generated from a modular latent factor model with $m=32$ and $p=128$.}
\label{tab:syn_results_m32}
\end{table*}

\begin{table}[ht]
\begin{center}
\begin{tabular}{lccccc}
\toprule
Method     &   $w=12$    &   $w=24$   &   $w=48$   &   $w=96$   \\
\midrule
LW         & 424.9 $\pm$ 40.1 & 384.7 $\pm$ 100.3 & 372.4 $\pm$ 54.6 & 465.6 $\pm$ 21.9 \\
FA         & - & - & 900.3 $\pm$ 189.0 & 355.7 $\pm$ 15.2 \\
Sparse PCA & 507.7 $\pm$ 71.5 & 382.9 $\pm$ 66.7 & 301.1 $\pm$ 31.2 & 266.5 $\pm$ 10.8 \\
Linear CorEx & 458.1 $\pm$ 47.6 & 330.1 $\pm$ 31.9 & 279.7 $\pm$ 14.6 & 262.2 $\pm$ 8.5 \\
GLASSO     & 473.3 $\pm$ 65.5 & 400.3 $\pm$ 54.8 & 304.8 $\pm$ 33.6 & 272.4 $\pm$ 18.0 \\
LVGLASSO   & 430.6 $\pm$ 23.6 & 372.7 $\pm$ 59.0 & 289.3 $\pm$ 22.1 & 282.8 $\pm$ 9.7 \\
TVGL       & 335.6 $\pm$ 24.8 & 298.3 $\pm$ 46.0 & 260.5 $\pm$ 32.6 & 243.3 $\pm$ 6.9 \\
LTGL       & 329.3 $\pm$ 18.1 & 308.5 $\pm$ 49.1 & \textit{252.0 $\pm$ 27.9} & \textbf{231.8 $\pm$ 9.8}\\
T-CorEx    & \textbf{313.5 $\pm$ 16.4} & \textbf{269.5 $\pm$ 16.1} & \textbf{246.1 $\pm$ 12.9} & 244.0 $\pm$ 4.8 \\
\bottomrule
\end{tabular}
\end{center}
\caption{Time-averaged negative log-likelihood of the estimates produced by different methods on stock market data, shown in mean $\pm$ standard deviation format. The means and standard deviations are computed for 10 random train/validation/test splits.}
\label{tab:stock-day}
\end{table}

\section{Experimental Details}\label{sec:app-experimental-details}
In this subsection we describe the hyperparameter grids we used in quantitative experiments.
Tables \ref{tab:sudden-hypers}, \ref{tab:smooth-hypers}, and  \ref{tab:stock-hypers} describe these grids for the sudden change, smooth change, and stock market experiments respectively.
Note that the excluded baselines (beside T-CorEx-simple and T-CorEx-no-reg) have no hyperparameters.
In our experiments all methods are trained for 500 iterations.
We made sure that increasing it does not improve the results of any baseline.
Also, we checked that the best values of hyperparameters are not the corner values.
In the sudden and smooth change experiments we know the ground truth value of $m$ (the number of latent factors).
Therefore, we do not search the best value of $m$ for FA, sparse PCA, linear CorEx, and T-CorEx baselines.
This way we compare the modeling performances of baselines, rather than their sensitivity to hyperparameters.
\begin{table}{b}
    \centering
    
    \begin{tabular}{ll}
    \toprule
    Method & Hyperparameters \\
    \midrule
    Sparse PCA & \begin{tabular}{ll}
          alpha &  [0.1, 0.3, 1, 3, 10, 30]
    \end{tabular}\\
    \midrule
    GLASSO & \begin{tabular}{ll}
          lamb &  [0.01, 0.03, 0.1, 0.3, 1, 3]
    \end{tabular}\\
    \midrule
    LVGLASSO & \begin{tabular}{ll}
         alpha & [0.03, 0.1, 0.3, 1, 3, 10]\\
         tau & [1.0, 3.0, 10.0, 30, 100, 300]\\
    \end{tabular}\\
    \midrule
    TVGL & \begin{tabular}{ll}
         lamb & [0.01, 0.03, 0.1, 0.3, 1, 3]\\
         beta & [0.03, 0.1, 0.3, 1, 3, 10] \\
         indexOfPenalty & [1, 2, 3]
    \end{tabular}\\
    \midrule
    LTGL & \begin{tabular}{ll}
         alpha & [0.3, 1, 3, 10]\\
         tau & [10, 30, 100, 300, 1e3] \\
         beta & [1, 3, 10, 30, 100] \\
         psi & [l1, l2, Laplacian]\\
         eta & [3, 10, 30]\\
         phi & [l1, l2, Laplacian]\\
    \end{tabular}\\
    \midrule
    T-CorEx & \begin{tabular}{ll}
         l1 ($\lambda$) & [0.0, 0.03, 0.1, 0.3, 1, 3, 10]\\
         beta & [1e-9, 0.1, 0.3, 0.4, 0.5, 0.6, 0.7]
    \end{tabular}\\
    \bottomrule
    \end{tabular}
    \caption{Hyperparameter grids in the \textit{sudden change} experiment.}
    \label{tab:sudden-hypers}
\end{table}
\begin{table}
    \centering
    \begin{tabular}{ll}
    \toprule
    Method & Hyperparameters \\
    \midrule
    Sparse PCA & \begin{tabular}{ll}
          alpha &  [0.1, 0.3, 1, 3, 10, 30]
    \end{tabular}\\
    \midrule
    GLASSO & \begin{tabular}{ll}
          lamb &  [0.01, 0.03, 0.1, 0.3, 1, 3]
    \end{tabular}\\
    \midrule
    LVGLASSO & \begin{tabular}{ll}
         alpha & [0.03, 0.1, 0.3, 1, 3, 10]\\
         tau & [1, 3, 10, 30, 100, 300]
    \end{tabular}\\
    \midrule
    TVGL & \begin{tabular}{ll}
         lamb & [0.01, 0.03, 0.1, 0.3, 1, 3]\\
         beta & [0.03, 0.1, 0.3, 1, 3, 10] \\
         indexOfPenalty & [1, 2, 3]
    \end{tabular}\\
    \midrule
    LTGL & \begin{tabular}{ll}
         alpha & [3, 10, 30, 1e2]\\
         tau & [30, 100, 300, 1e3, 3e3] \\
         beta & [1, 3, 10, 30, 100] \\
         psi & [l1, l2, Laplacian]\\
         eta & [1, 3, 10]\\
         phi & [l1, l2, Laplacian]
    \end{tabular}\\
    \midrule
    T-CorEx & \begin{tabular}{ll}
         l2 ($\lambda$) & [0, 0.1, 0.3, 1, 3, 10, 30, 100, 300]\\ 
         beta & [1e-9, 0.1, 0.3, 0.4, 0.5, 0.6, 0.7]
    \end{tabular}\\
    \bottomrule
    \end{tabular}
    \caption{Hyperparameter grids in the \textit{smooth change} experiment.}
    \label{tab:smooth-hypers}
\end{table}
\begin{table*}[!t]
    \centering
    \begin{tabular}{ll}
    \toprule
    Method & Hyperparameters \\
    \midrule
    FA & \begin{tabular}{ll}
        n\_components & [16, 32, 64, 128]
    \end{tabular}\\
    \midrule
    Sparse PCA & \begin{tabular}{ll}
          alpha &  [0.1, 0.3, 1, 3, 10, 30]\\
          n\_components & [16, 32, 64, 128]
    \end{tabular}\\
    \midrule
    Linear CorEx & \begin{tabular}{ll}
          n\_hidden ($m$) & [16, 32, 64, 128]
    \end{tabular}\\
    \midrule
    GLASSO & \begin{tabular}{ll}
          lamb &  [0.01, 0.03, 0.1, 0.3, 1, 3]
    \end{tabular}\\
    \midrule
    LVGLASSO & \begin{tabular}{ll}
         alpha & [0.03, 0.1, 0.3, 1, 3, 10]\\
         tau & [1, 3, 10, 30, 100, 300]
    \end{tabular}\\
    \midrule
    TVGL & \begin{tabular}{ll}
         lamb & [0.01, 0.03, 0.1, 0.3, 1, 3]\\
         beta & [0.03, 0.1, 0.3, 1, 3, 10] \\
         indexOfPenalty & [1, 2]
    \end{tabular}\\
    \midrule
    LTGL & \begin{tabular}{ll}
         alpha & [0.3, 1, 3, 10]\\
         tau & [30, 100, 300, 1e3] \\
         beta & [10, 30, 100] \\
         psi & [l1, l2, Laplacian]\\
         eta & [0.3, 1, 3]\\
         phi & [l1, l2, Laplacian]
    \end{tabular}\\
    \midrule
    T-CorEx & \begin{tabular}{ll}
         l1 ($\lambda$) & [0, 0.03, 0.1, 0.3, 1, 3, 10] \\
         beta & [1e-9, 0.1, 0.2, 0.3, \ldots, 0.9]\\
         n\_hidden ($m$) & [16, 32, 64, 128]
    \end{tabular}\\
    \bottomrule
    \end{tabular}
    \caption{Hyperparameter grids in the \textit{stock market} experiment.}
    \label{tab:stock-hypers}
\end{table*}
\vfill

\end{document}